\documentclass[11pt,a4paper]{article}
\usepackage[hyperref]{emnlp2020}
\usepackage{times}
\usepackage{latexsym}
\usepackage{multirow,booktabs,makecell}
\usepackage{graphicx}
\usepackage{enumitem}
\usepackage{subcaption}
\usepackage{xcolor,colortbl}
\usepackage{siunitx}
\usepackage{booktabs}
\usepackage[version=4]{mhchem}
\usepackage{collcell}
\usepackage{adjustbox}
\usepackage{microtype}

\aclfinalcopy % Uncomment this line for the final submission
\def\aclpaperid{3504} %  Enter the acl Paper ID here

\title{Towards Understanding Sample Variance in Visually Grounded\\ Language Generation: Evaluations and Observations}

\author{Wanrong Zhu\textsuperscript{\dag}, Xin Eric Wang\textsuperscript{\ddag}, Pradyumna Narayana\textsuperscript{*}, \\
\textbf{Kazoo Sone\textsuperscript{*}, Sugato Basu\textsuperscript{*}, William Yang Wang\textsuperscript{\dag}} \\
\textsuperscript{\dag}UC Santa Barbara, 
\textsuperscript{\ddag}UC Santa Cruz, 
\textsuperscript{*}Google \\
\texttt{\small \{wanrongzhu,william\}@cs.ucsb.edu}, 
\texttt{\small xwang366@ucsc.edu},
\texttt{\small \{pradyn,sone,sugato\}@google.com}
}

\date{}

\begin{document}
\maketitle

%=========================ABSTRACT==================================================
\begin{abstract}

A major challenge in visually grounded language generation is to build robust benchmark datasets and models that can generalize well in real-world settings.
To do this, 
it is critical to ensure that our evaluation protocols are correct, and benchmarks are reliable.  
%However, a primary focus of the research community has been concentrating on pushing the ``state-of-the-art'' numbers higher on leaderboards, rather than understanding the nature of these numbers and tasks. 
In this work, we set forth to design a set of experiments to understand an important but often ignored problem in visually grounded language generation: given that humans have different utilities and visual attention, how will the sample variance in multi-reference datasets affect the models' performance? Empirically, 
we study several multi-reference datasets and corresponding vision-and-language tasks.
We show that it is of paramount importance to report variance in experiments; that human-generated references could vary drastically in different datasets/tasks, revealing the nature of each task; that metric-wise, CIDEr has shown systematically larger variances than others.
Our evaluations on reference-per-instance shed light on the design of reliable datasets in the future.
\end{abstract}

%=========================INTRODUCTION==================================================
\section{Introduction}

\begin{figure}[t]
\centering
\includegraphics[width=\linewidth]{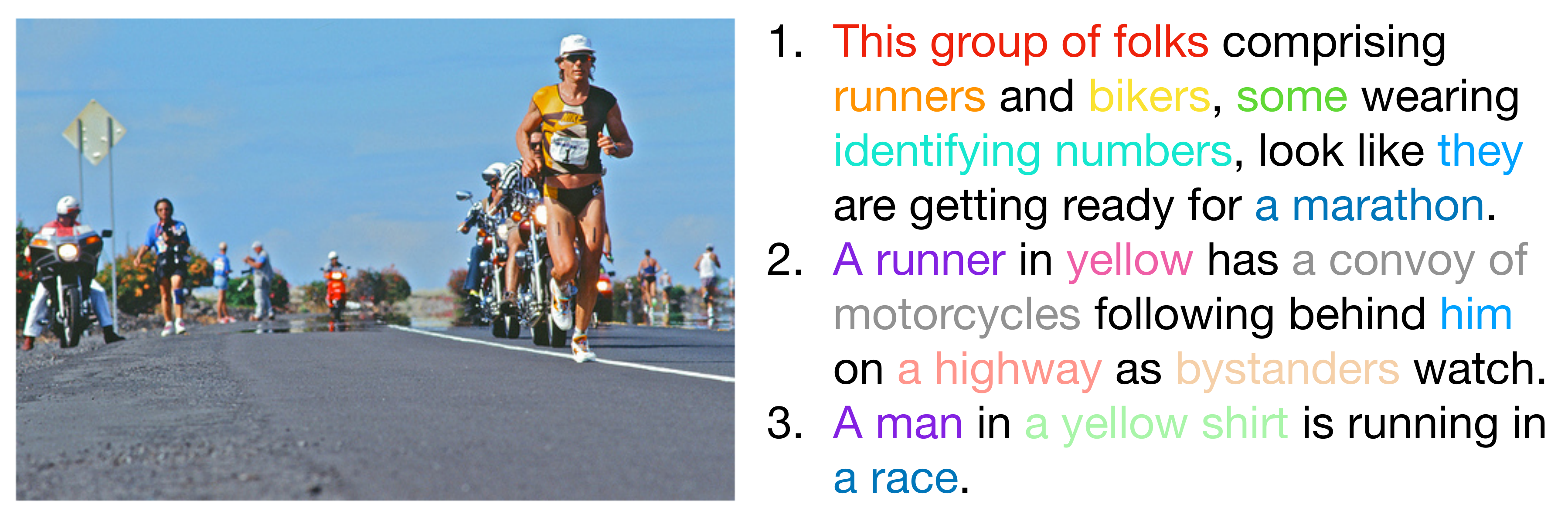}
\caption{An image with three parallel captions from the Flickr30k dataset. Words in the same colors refer to the same objects.}
%} %Parallel captions may be describing different content of the image and have distinct sentence structure and length.}
\label{fig:variance_example}
\end{figure}

Natural Language Generation (NLG) is a challenging problem in Natural Language Processing (NLP)---the complex nature of NLG tasks arise particularly in the output space. In contrast to text classification or regression problems with finite output space, generation could be seen as a combinatorial optimization problem, where we often have exponentially many options $|V|^{\ell}$ (here $|V|$ is the size of the vocabulary and $\ell$ is the sentence length). With the advances of both Computer Vision and NLP techniques in deep learning,
there have been growing interests in visually grounded NLG tasks, such as image captioning~\citep{flickr8k, flickr30k, coco, cider}, video captioning~\citep{xu2016msr, wang2019vatex, chen-dolan-2011-collecting} and visual storytelling~\citep{huang2016visual}. For example, Figure~\ref{fig:variance_example} shows an example of image captioning from the popular Flickr30k dataset.  

In this paper, instead of crunching numbers 
and modifying model architectural designs to achieve new ``state-of-the-art'' results on leaderboards, we focus on re-assessing the current practices in visually grounded language generation research, including problems, datasets, evaluations, and tasks, from the sample variance angle. 
%From a Statistical Machine Learning perspective, an intriguing scientific research question is: given only a few references for each visual instance, how can learners approximate the distribution in this vast search space? From Cognitive Psychology, 
Given the differences in annotators' utility function and human visual attention models, how could the sample variance in captions teach us building robust and reliable visually grounded language generation agents?

More specifically, we empirically investigate the variance among the multiple parallel references in different datasets, and its effect on the training performance and evaluation result of corresponding tasks. We further study the number of references per visual instance, and how it affects the training and testing performance. 
A simple search in ACL Anthology and CVF Open Access Site shows that 58 out of 60 papers on vision-based text generation do not report variance in experimental results, while they often claim that their methods outperform previous state-of-the-art. Our evaluation
suggests that the variance cannot be ignored and must be reported, and that CIDEr~\citep{cider} has shown higher variance than other metrics. Finally,
introducing more training visual instances in the image and video captioning task on MS COCO and VATEX results in better performance on automatic metrics, while the visual storytelling task in VIST favors more references in the training set. For future dataset collection, we recommend the inclusion of more references when each reference is distinctive and complicated.
%\ww{Again, I think the variability affects a lot of different things and touches many issues. We don't want to be too narrow in the scope of our paper.}
% What is the problem we studied?

% %===========================================================================
\section{Research Questions and Settings}

To understand sample variance, we conduct a series of experiments on multiple visually grounded NLG datasets, aiming to answer the following questions:
\begin{enumerate}
    \item \emph{How different are the text references from their parallel pairs?}
    \item \emph{How greatly do different selections of references during either training or testing affect the final evaluation results?}
    \item \emph{To train a more reliable model, shall we collect more visual instances with limited references or more parallel references for each instance given a fixed budget?}
\end{enumerate}

We focus on multi-reference visually grounded NLG tasks where each visual instance is paired with multiple parallel text references. Below we describe the datasets we investigate into, the models used for training, and the metrics for evaluation.

%-----------------------------------------
\paragraph{Datasets}
Seven commonly used datasets in Table~\ref{tab:dataset_details} are considered:  Flickr8k~\citep{flickr8k}, Flickr30k~\citep{flickr30k}, MS COCO~\citep{coco}, PASCAL-50S~\citep{cider}, VATEX\_en (English), VATEX\_cn (Chinese)~\citep{wang2019vatex}, and VIST~\citep{huang2016visual}, covering the tasks of image captioning, video captioning, and visual storytelling.

%-----------------------------------------
\paragraph{Models}
We apply an implementation\footnote{\href{https://github.com/sgrvinod/a-PyTorch-Tutorial-to-Image-Captioning}{https://github.com/sgrvinod/a-PyTorch-Tutorial-to-Image-Captioning}} of ~\citet{xu2015show} for image captioning. 
We implement the Enc-Dec baseline model proposed by ~\citet{wang2019vatex} for video captioning.
For visual storytelling, we use the AREL model\footnote{\href{https://github.com/eric-xw/AREL}{https://github.com/eric-xw/AREL}} proposed by~\citet{wang2018no}.

%-----------------------------------------
\paragraph{Metrics}
We utilize six automatic metrics for natural language generation to evaluate the quality of the generated text, including BLEU~\citep{BLEU}, ROUGE~\citep{rouge}, METEOR~\citep{meteor}, CIDEr~\citep{cider}, SPICE~\citep{spice} and the most recent BERTScore~\citep{bert-score} that is based on the pretrained BERT model.

We use nlg-eval\footnote{\href{https://github.com/Maluuba/nlg-eval}{https://github.com/Maluuba/nlg-eval}} ~\citep{sharma2017nlgeval} for the calculation of BLEU, METEOR, ROUGE\_L and CIDEr. 
Note that we applied a patch\footnote{\href{https://github.com/vrama91/coco-caption}{https://github.com/vrama91/coco-caption}} and choose to use IDF from the MSCOCO Vaildation Dataset when calculating consensus CIDEr score for each dataset. We use the authors' releases for SPICE\footnote{\href{https://github.com/peteanderson80/SPICE}{https://github.com/peteanderson80/SPICE}} and BERTScore\footnote{\href{https://github.com/Tiiiger/bert\_score}{https://github.com/Tiiiger/bert\_score}}. BERTScore has been rescaled with baseline scores.

\begin{table}
\small
\setlength{\tabcolsep}{3pt}
\begin{center}
\resizebox{\columnwidth}{!}{
\begin{tabular}{l l S[table-format=2.0] S[table-format=2.2] S[table-format=2.0] S[table-format=1.0] S[table-format=1.0]}

\cmidrule[\heavyrulewidth]{1-7}
\textbf{Task}                       & \textbf{Dataset}  & \textbf{\#ref}  & \textbf{\#len} &  \textbf{\#train} & \textbf{\#val}& \textbf{\#test}  \\ \cmidrule{1-7}
% Task                       & Dataset & \#ref  & \#len &  \#train & \#val& \#test  \\ \cmidrule{1-7}
\multirow{4}{*}{Image Captioning}   & Flickr8k          & 5             &  11.8     & 6k              & 1k            & 1k  \\ \cmidrule{2-7}
                                    & Flickr30k         & 5             &  12.3     & 29k              & 1k            & 1k \\ \cmidrule{2-7}
                                    & MS COCO'14        & 5             &   10.5     & 83k              & 5k            & 5k \\ \cmidrule{2-7}
                                    & PASCAL-50S        & 50            &   8.8     & \textemdash                 & \textemdash             & 1k \\ \cmidrule{1-7}
\multirow{2}{*}{Video Captioning}   & VATEX\_en        & 10             &   15.2   &  26k                &  3k            & 6k \\ \cmidrule{2-7}
                                    & VATEX\_cn        & 10             &   14.0   &  26k                &  3k            & 6k \\ \cmidrule{1-7}
Visual Storytelling                 & VIST              & 5             &   56.8     & 8k                & 1k            & 1k   \\ 
\cmidrule[\heavyrulewidth]{1-7}
\end{tabular}
}
\end{center}
\caption{Dataset statistics. \textit{\#ref} is the number of parallel references per visual instance; \textit{\#len} is the average reference length; \textit{\#train, \#val}, and \textit{\#test} are the number of visual instances of training, validation, and test sets. }
\label{tab:dataset_details}
\end{table}

%===========================================================================
%-----------------------------------------
% \section{Reference Sample Variance within Datasets}
\section{Reference Variance within Datasets}
\label{sec:variance_within_dataset}

\begin{table*}[h!]
% \small
\sisetup{table-figures-uncertainty=1}
\begin{center}
\resizebox{\linewidth}{!}{
\begin{tabular}{l l 
>{\collectcell\num}r<{\endcollectcell}
  @{${}\pm{}$}
  >{\collectcell\num}r<{\endcollectcell}
  >{\collectcell\num}r<{\endcollectcell}
  @{${}\pm{}$}
  >{\collectcell\num}r<{\endcollectcell}
  >{\collectcell\num}r<{\endcollectcell}
  @{${}\pm{}$}
  >{\collectcell\num}r<{\endcollectcell}
  >{\collectcell\num}r<{\endcollectcell}
  @{${}\pm{}$}
  >{\collectcell\num}r<{\endcollectcell}
  >{\collectcell\num}r<{\endcollectcell}
  @{${}\pm{}$}
  >{\collectcell\num}r<{\endcollectcell}
 >{\collectcell\num}r<{\endcollectcell}
  @{${}\pm{}$}
  >{\collectcell\num}r<{\endcollectcell}
  }
% \begin{tabular}{@{} ll *{6}{S} @{}}
% \begin{tabular}{l l S[table-format=2.2] S[table-format=2.2] S[table-format=2.2] S[table-format=2.2] S[table-format=2.2] S[table-format=2.2] }
\cmidrule[\heavyrulewidth]{1-14}
% \textbf{Dataset}  & $\overline{len}$    &  $\overline{BLEU}$    & $\overline{ROUGE_L}$  & $\overline{METEOR}$ & $\overline{CIDER}$\\ \cmidrule{1-6}
\textbf{Task}    & \textbf{Dataset}    & \multicolumn{2}{c}{\textbf{BLEU}}     & \multicolumn{2}{c}{\textbf{METEOR}}   & \multicolumn{2}{c}{\textbf{ROUGE\_L}} & \multicolumn{2}{c}{\textbf{CIDEr}} & \multicolumn{2}{c}{\textbf{SPICE}}    & \multicolumn{2}{c}{\textbf{BERTScore}}  \\ \cmidrule{1-14}
\multirow{4}{*}{Image Captioning} & Flickr8k            &	35.05 & 12.63	&	26.72 & 7.65	&	49.85 & 11.93	&	85.23 & 57.53	&	23.22 & 10.00	&	58.40 & 10.76 \\ \cmidrule{2-14} % \cellcolor{red!20}
& Flickr30k           &	32.22 & 11.98	&	23.98 & 7.22	&	45.15 & 11.75	&	65.24 & 50.31	&	19.46 & 8.63	&	52.77 & 11.14 \\ \cmidrule{2-14}
& MS COCO'14          &	33.52 & 12.05	&	24.70 & 6.88	&	46.60 & 11.06	&	86.09 & 53.39	&	21.11 & 8.46	&	54.40 & 10.98 \\ \cmidrule{2-14}
& PASCAL-50S          &	33.60 & 8.88	&	26.54 & 5.59	&	50.18 & 9.08	&	89.35 & 41.25	&	23.04 & 6.47	&	57.26 & 9.00 \\ \cmidrule{1-14}
\multirow{2}{*}{Video Captioning}   & VATEX\_en          &	30.64 & 7.87	&	22.07 & 4.48	&	40.65 & 7.41	&	64.45 & 34.46	&	18.28 & 5.65	&	48.99 & 8.06 \\ \cmidrule{2-14}
& VATEX\_cn           &	25.08 & 6.52	&	25.63 & 3.99	&	40.40 & 6.21	&	87.28 & 25.89	&	31.59 & 5.22	&	50.40 & 7.05 \\  \cmidrule{1-14}
Visual Storytelling & VIST                &	18.42 & 4.37	&	12.53 & 2.23	&	20.54 & 3.41   	&	11.46 & 9.13	&	8.95 & 2.81    &	15.46 & 6.58 \\
\cmidrule[\heavyrulewidth]{1-14}
\end{tabular}
}
\end{center}
\caption{The mean and standard deviation of consensus score for each metric on all the datasets.}
\label{tab:dataset_inner_variance}
\end{table*}

\begin{figure*}[t!]
    \centering
    \begin{subfigure}[t]{0.33\textwidth}
        \centering
        \includegraphics[width=5.2cm]{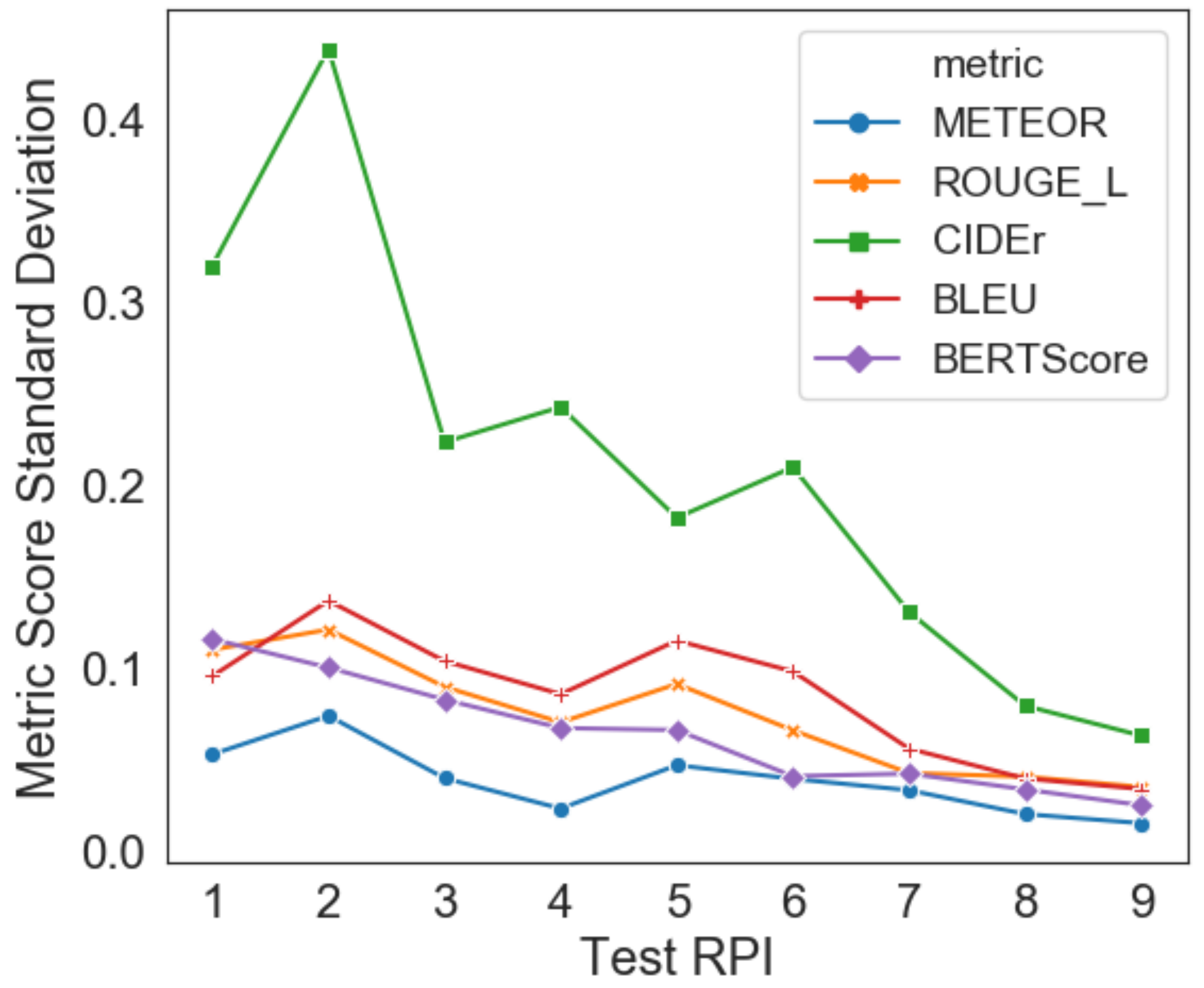}
        \caption{Score deviation on VATEX\_en.\label{fig:exp3_vatex_en_std}}
    \end{subfigure}%
    ~ 
    \begin{subfigure}[t]{0.33\textwidth}
        \centering
        \includegraphics[width=5.2cm]{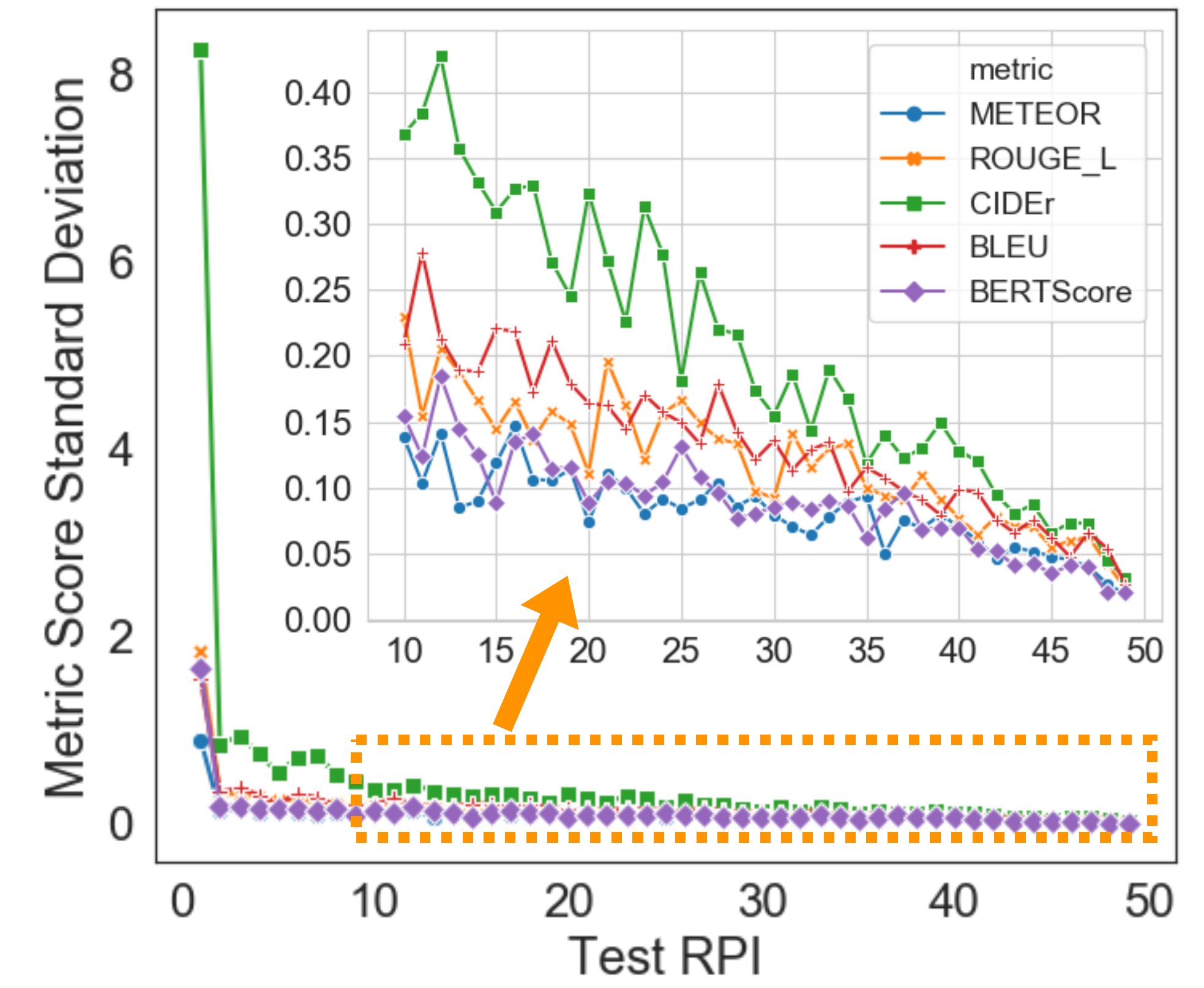}
        \caption{Score deviation on PASCAL50S.\label{fig:exp3_pascal_cn_std}}
    \end{subfigure}%
    ~ 
    \begin{subfigure}[t]{0.33\textwidth}
        \centering
        \includegraphics[width=5.2cm]{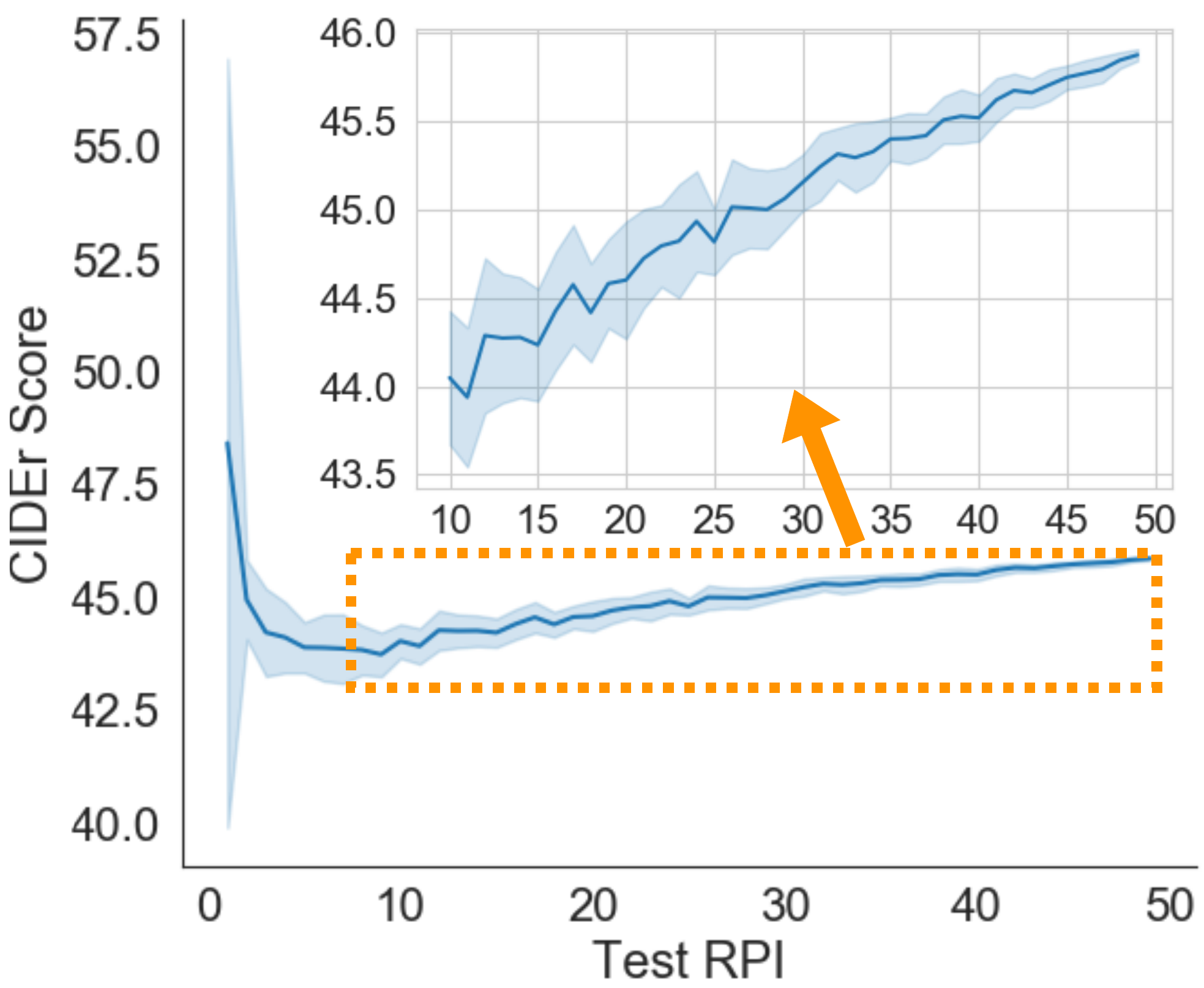}
        \caption{CIDEr score on PASCAL50S.\label{fig:exp3_pascal_cider}}
    \end{subfigure}%
\caption{Effect of varying testing RPI for evaluation. \label{fig:test_cpi}}
\end{figure*}

\begin{table}
\begin{adjustbox}{width=\linewidth,center}
\begin{tabular}{l | r }
\cmidrule[\heavyrulewidth]{1-2}
\textbf{Reference}  & \textbf{CIDEr} \\ \cmidrule[\heavyrulewidth]{1-2}
A man riding an elephant in a river. & 225    \\ 
A man in a brown shirt rides an elephant into the water. & 227     \\
A man rides an elephant into a river. & 266     \\ 
A man riding an elephant into some water of a creek. & 271     \\ 
Man riding an elephant into water surrounded by forest. & 277     \\  
\cmidrule[\heavyrulewidth]{1-2}
There are many taxi cabs on the road & 4 \\
Heavy city traffic all going in one direction & 26 \\
Many cars stuck in traffic on a high way & 28 \\
This shot is of a crowded highway full of traffic & 28 \\
A city street with lots of traffic and lined with buildings & 35 \\
\cmidrule[\heavyrulewidth]{1-2}
\end{tabular}
\end{adjustbox}
\caption{Two group of references from MSCOCO dataset and the CIDEr score for each reference within their group. The consensus CIDEr score for the two groups of references are 253.2 and 24.2 respectively.}
\label{tab:cider_example}
\end{table}

In this section, we examine the sample variance among text references within seven visually grounded NLG datasets. To quantify the sample variance, we define a consensus score $c$ among $n$ parallel references $R = \{r_i\}_{i=1}^n$ (where $r_i$ is the $i$-th text reference) for each visual instance:
\begin{equation}
    c = \frac{1}{n} \sum_{i=1}^n metric (r_i, R \backslash \{r_i\})
\label{eq:eval}
\end{equation}
where $metric$ can be any metric in the above section. The consensus score represents the agreement among the parallel references for the same visual instance. 
Since the number of parallel references varies across datasets, we randomly sample 5 parallel references per instance (the minimum $n$ all datasets used) for a fair comparison. For datasets with more than 5 parallel references per instance, we repeat 10 times and take the average.

Table \ref{tab:dataset_inner_variance} shows the evaluation results.
Noticeably, the datasets for the same task have similar consensus BERTScore, which is embedding-based~\citep{kilickaya-etal-2017-evaluating}. Image captioning datasets score the highest on BERTScore consensus, video captioning datasets rank the second, while VIST for visual storytelling has the lowest consensus BERTScore.
The descending consensus BERTScore order coincides with task difficulties. Video captioning is more complicated than image captioning due to its dynamic nature.
Visual storytelling is even more challenging with the diverse and sophisticated stories in creative writing.
Having the lowest consensus scores on all metrics indicates that VIST is a very challenging dataset.
Moreover, we notice that CIDEr has the largest standard deviation (both absolutely and relatively) on consensus scores for all datasets.
This suggests that CIDEr might be unstable and sensitive to the selection of references.

\begin{figure*}[t]
    \centering
    \begin{subfigure}[t]{0.33\textwidth}
        \centering
        \includegraphics[width=5.2cm]{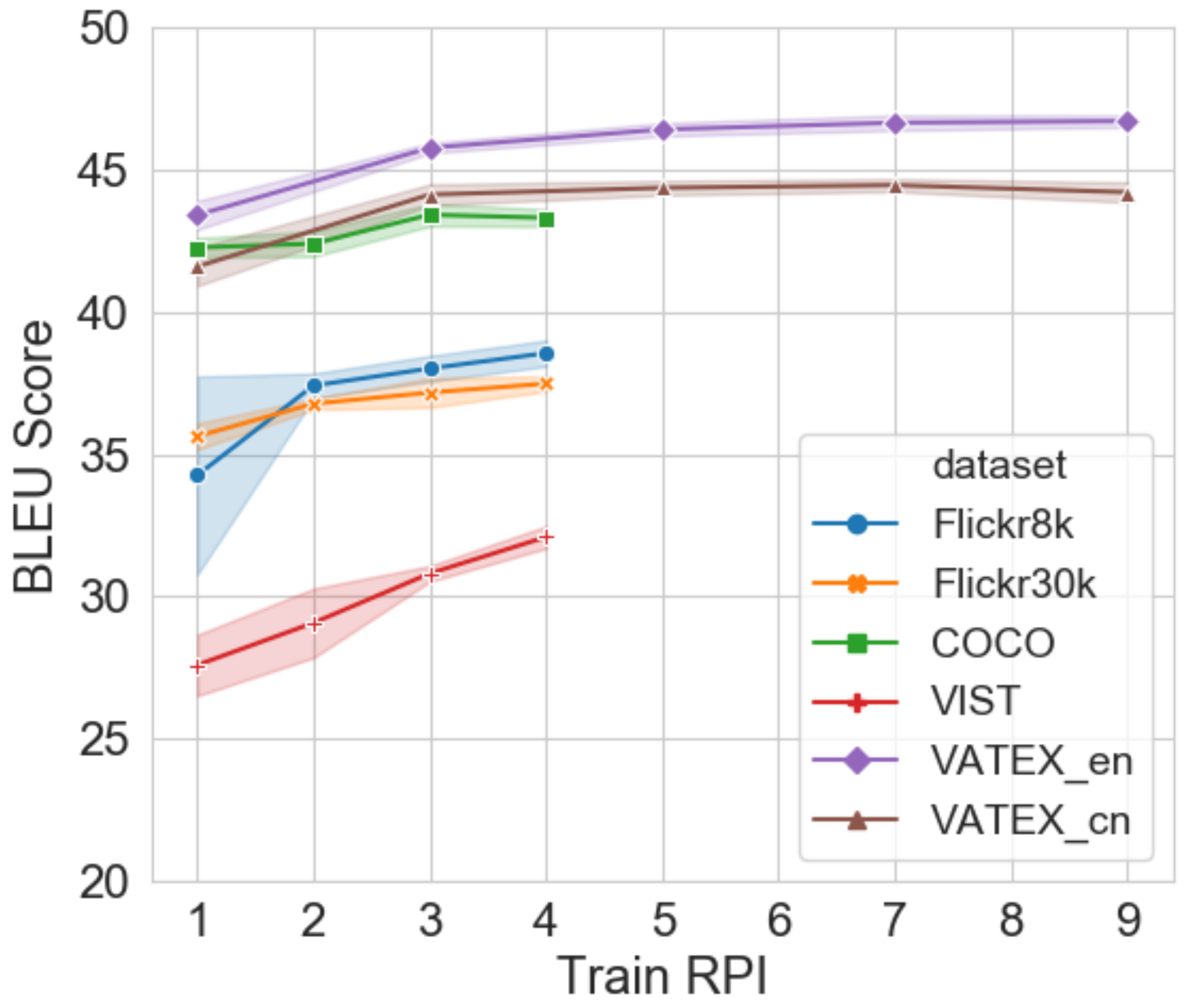}
        \caption{BLEU Score\label{fig:exp1_bleu}}
    \end{subfigure}%
    ~ 
    \begin{subfigure}[t]{0.33\textwidth}
        \centering
        \includegraphics[width=5.2cm]{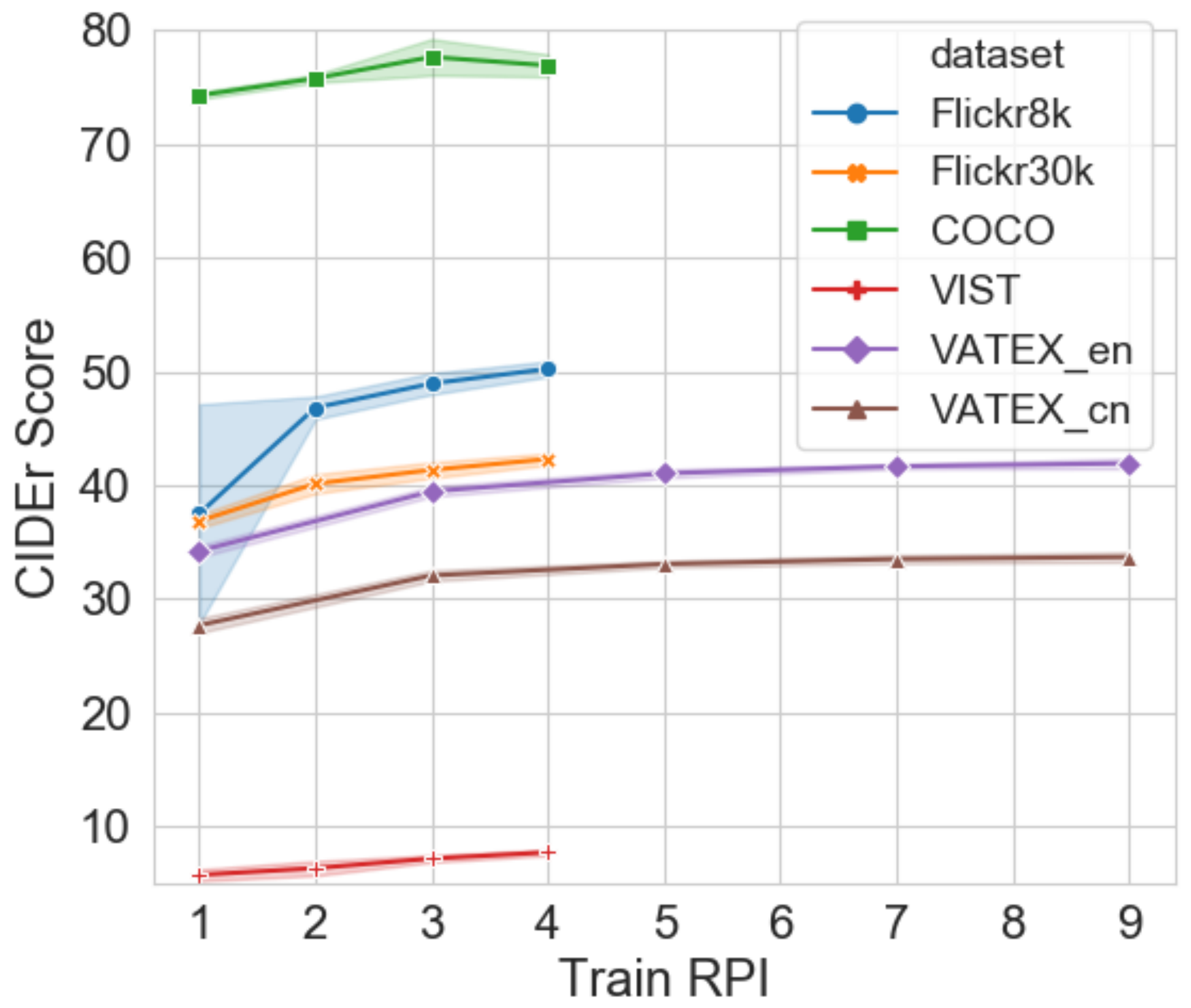}
        \caption{CIDEr Score\label{fig:exp1_cider}}
    \end{subfigure}%
    ~ 
    \begin{subfigure}[t]{0.33\textwidth}
        \centering
        \includegraphics[width=5.2cm]{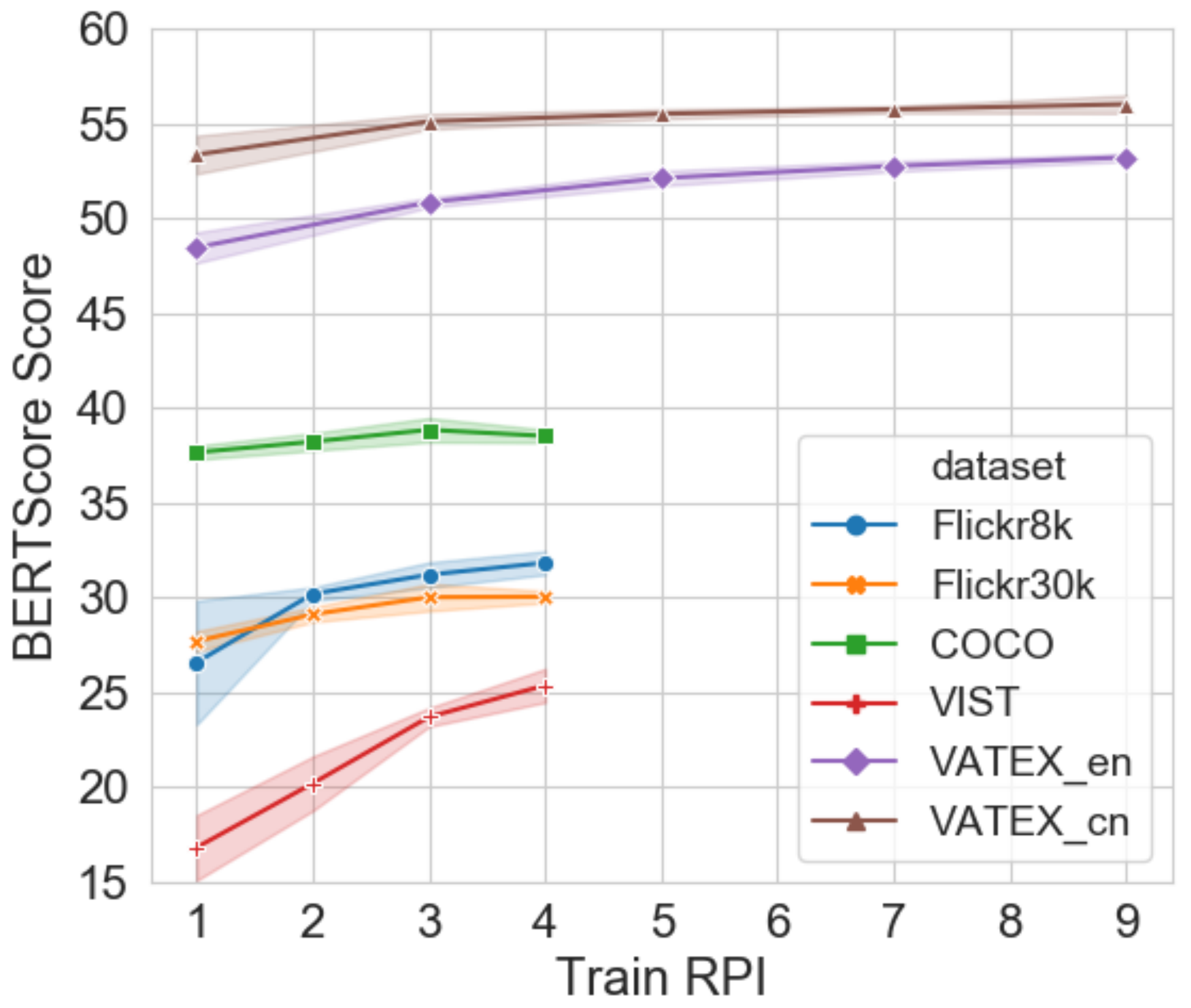}
        \caption{BERTScore\label{fig:exp1_bert}}
    \end{subfigure}%
\caption{Performance when trained with varying training RPI on all the visual instances of the training set.
\label{fig:train_cpi}
}
\end{figure*}

Table \ref{tab:cider_example} takes a closer look at the high variance of the consensus CIDEr score. By definition, CIDEr score computes cosine similarity between the Term Frequency Inverse Document Frequency (TF-IDF) \citep{Robertson2004UnderstandingID} weighted n-grams. The reasons for the consensus CIDEr score to have high standard deviation are threefold:
(1) N-grams with similar meanings might have totally different TF-IDF weights. Therefore, the CIDEr score is sensitive to word selection and sentence structure. 
(2) Token frequency differs across datasets. The consensus CIDEr score in Table \ref{tab:dataset_inner_variance} is calculated on the sentence level. We follow previous work and use IDF from the MSCOCO validation set for reliable results. In the MSCOCO validation set, ‘man’, ‘elephant’, and ‘river’ have more exposure, while ‘traffic’ and ‘highway’ are less mentioned. As a result, the first group of references has a much higher consensus CIDEr score than the second group.
(3) Moreover, different from other metrics that scale from 0-1, the CIDEr score scales from 0-10. The enlarged scale also contributes to its salient variance.

%-----------------------------------------
% Evaluation Variance with Different Selections of References
% Effect of Reference Number on Evaluation Variance
\section{Effect of Sample Variance on Evaluation Performance}
\label{sec:rpi}

\begin{figure*}[t!]
    \centering
    \begin{subfigure}[t]{0.29\textwidth}
        \centering
        \includegraphics[width=4.6cm]{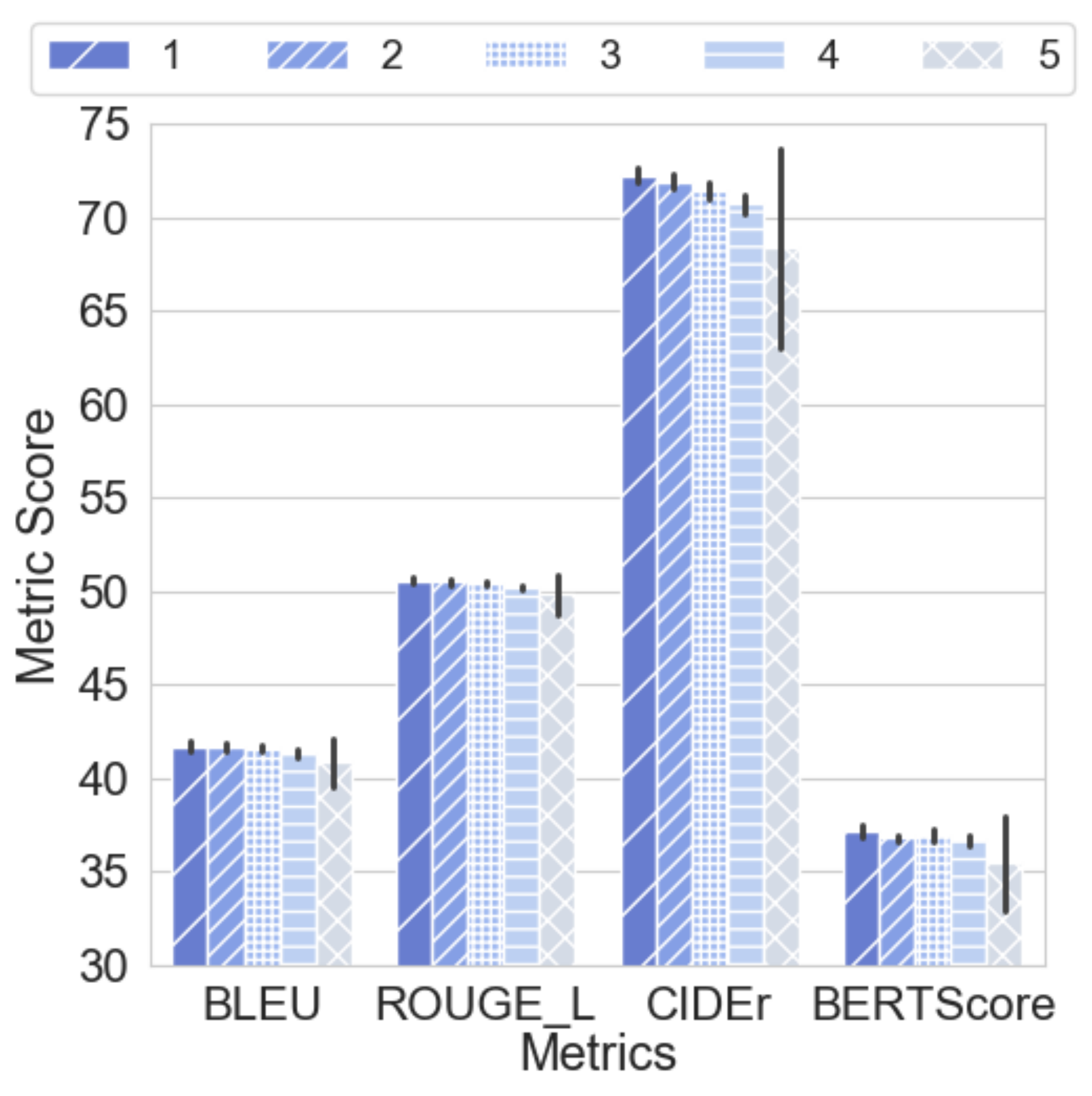}
        \caption{MS COCO\label{fig:exp2_coco}}
    \end{subfigure}%
    ~ 
    \begin{subfigure}[t]{0.4\textwidth}
        \centering
        \includegraphics[width=6.1cm]{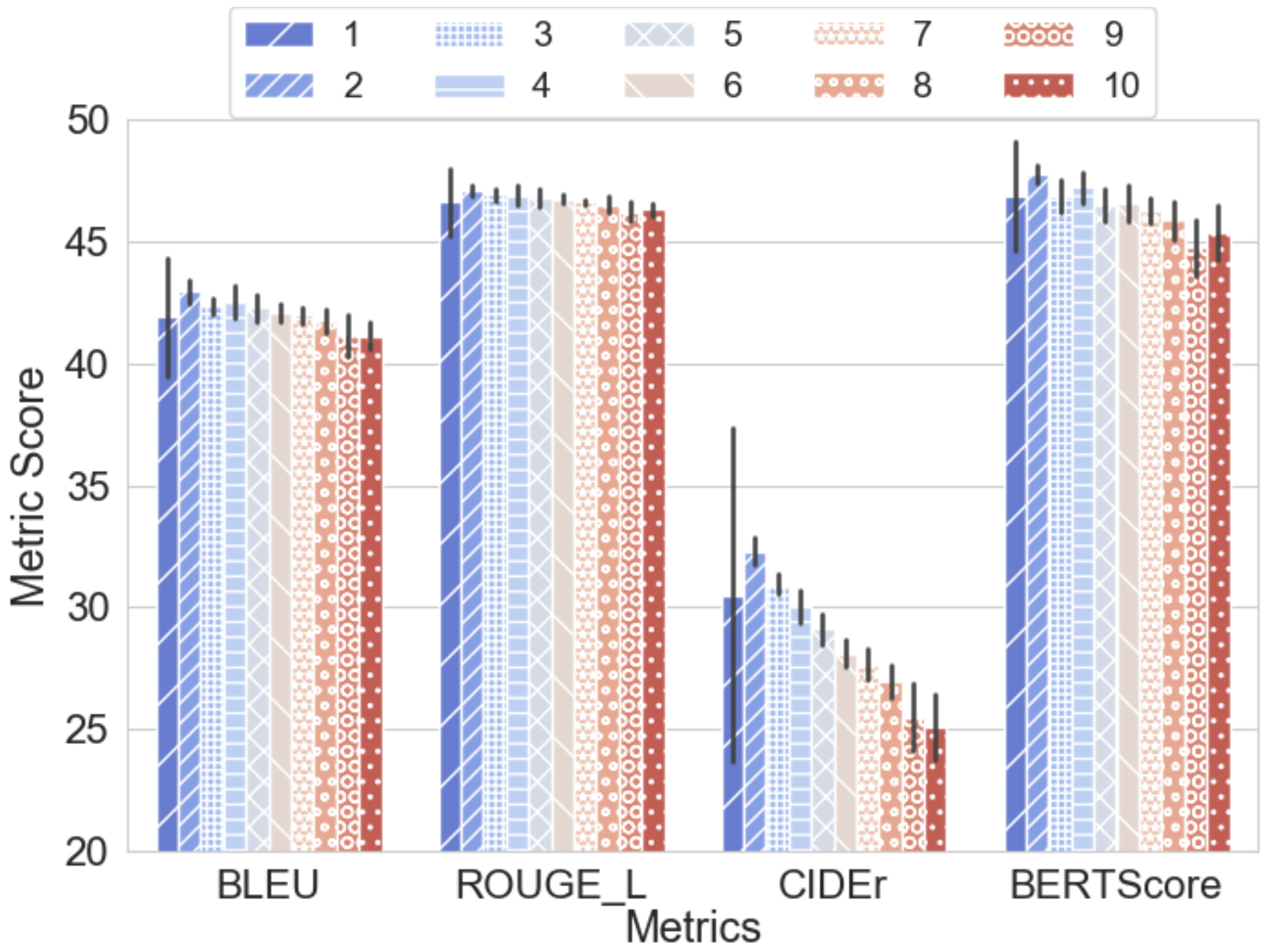}
        \caption{VATEX\_en\label{fig:exp2_vatex_en}}
    \end{subfigure}%
    ~ 
    \begin{subfigure}[t]{0.29\textwidth}
        \centering
        \includegraphics[width=4.6cm]{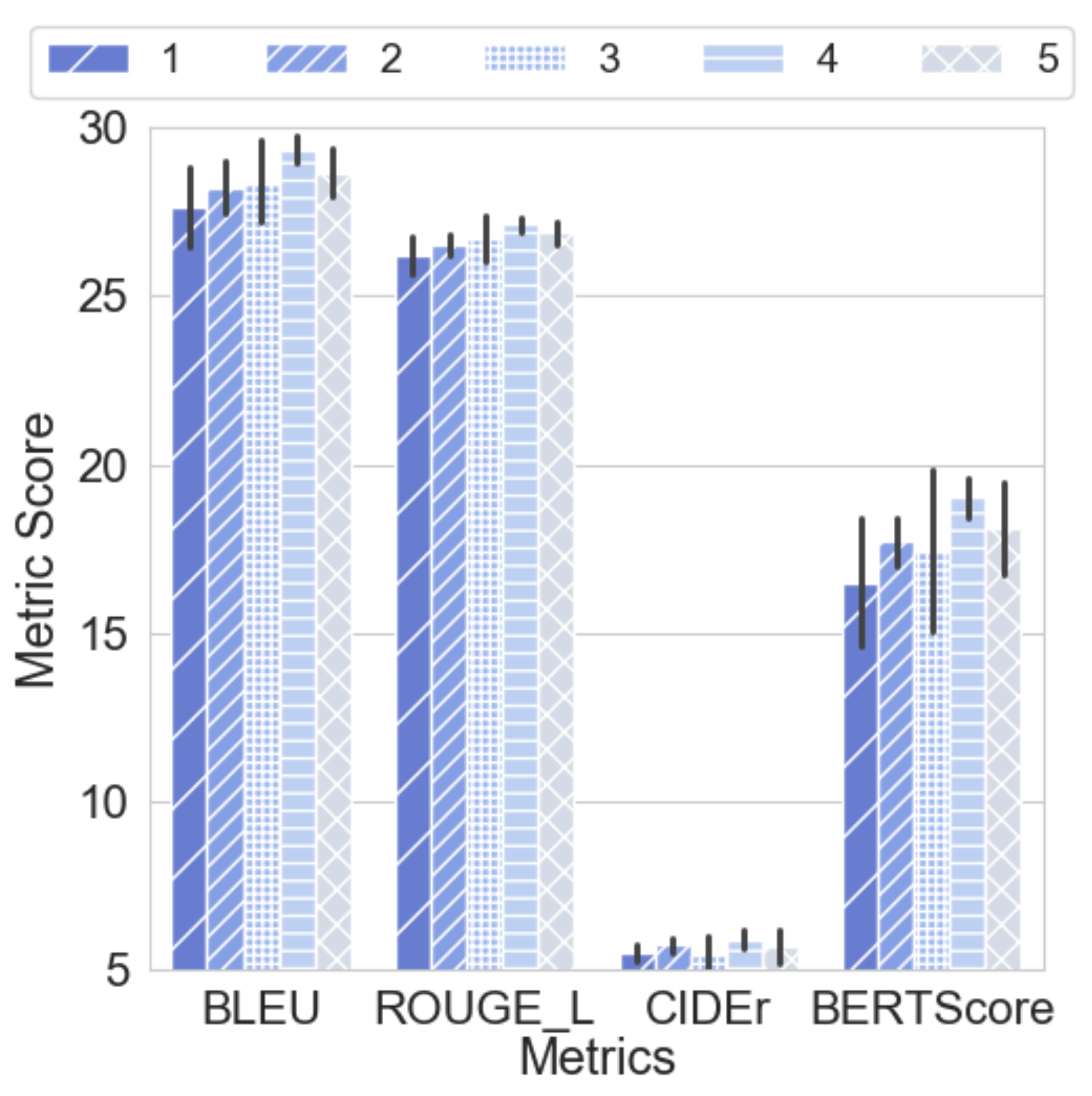}
        \caption{VIST\label{fig:exp2_vist}}
    \end{subfigure}%
\caption{
Performance when trained with varying training RPI on a fixed total number of visual-text sample pairs.
%Performance on MS COCO, VATEX\_en and VIST when models are trained with the same amount of samples, but different RPI.
Results on the captioning datasets COCO and VATEX\_en are in favor of more visual diversity, while the visual storytelling model benefits more from more parallel text references.
% The captioning performance on COCO and VATEX\_en get worse when more parallel references are introduced. On the contrary, the story generation performance on VIST improves with more parallel references during training.
\label{fig:ins_ref_balance}}
\end{figure*}

For visually grounded NLG tasks, models are trained on preset training samples and evaluated on preset testing samples, and then results are reported on leaderboards.
But would training or evaluating with different samples affect their performance? How reliable are those numbers?
In this section, we study to what extent the sample variance during either training or testing affects the evaluation results. 
For simplicity, the number of parallel References Per visual Instance used for training or testing is denoted by RPI.

\paragraph{Effect of Testing Sample Variance}
Previous studies on automatic metrics~\citep{cider, spice} show that more testing references lead to better evaluation accuracy.
Here we aim at examining the effect of using different references for testing. 
Given $n$ references per visual, we incrementally set the testing RPI as $1,2,...,n-1$, and randomly sample the testing references from all of the $n$ references. 
For each RPI, the random sampling and evaluation process is conducted for 20 times. The model is trained on the complete training set.

In  Figure~\ref{fig:test_cpi}, we demonstrate the experiments on PASCAL-50s for image captioning and VATEX\_en for video captioning, where the standard deviation of evaluation scores on those metrics are plotted over RPI.
For all metrics, the standard deviation shrinks as more references are employed for testing, indicating the evaluation bias caused by sample variance may be mitigated by introducing more parallel references.
% Special datasets with a larger amount of references such as PASCAL50S have been built for evaluating image captioning models. 
However, most of the existing datasets have far less than 50 references. For example, according to \citet{wang2019vatex}, 12 out of 15 datasets for video captioning have less than 3 parallel text references per video, but the variance on those metrics under 3 RPI is very high.
This casts doubt on the reliability of the model's performance.
For fairer model comparison, we hereby encourage researchers to (1) provide the evaluation set with more parallel references when collecting new datasets, and (2) report the variance of the model's metric scores as well when comparing to other models.
Noticeably, the variance of the model's performance on CIDEr is significantly larger than on other metrics, 
% especially when the testing RPI is small, 
which supplements the previous finding in Section~\ref{sec:variance_within_dataset} that CIDEr is very sensitive to the reference sample variance.

\paragraph{Effect of Training Sample Variance}
To investigate the effect of training sample variance, we train the models with different training RPI, from 1 to $n-1$. Similarly, we randomly sample the training references from $n$ references.
For each RPI, we repeat the random sampling and training process for 10 times on each dataset. The evaluation is conducted on the complete test set.

Figure \ref{fig:train_cpi} depicts the performance of BLEU, CIDEr and BERTScore on each dataset when the corresponding model is trained with different RPI.
While the performance on all datasets improves with the increase of training RPI, experimental results show salient variance on all metric scores when the amount of training data is insufficient, which indicates the selection of training samples will influence the final performance. Furthermore, VIST displays notable score deviation on all three metrics, which suggests visual storytelling to be sensitive to the selection of training data.

%-----------------------------------------
\section{More Visuals or More References?}
\label{sec:v_l_balance}

When collecting a new visually-grounded NLG dataset with a certain budget, there often exists a decision between collecting more visual instances v.s. more text references for each visual. How many parallel references do we need to train a reliable model for visual-grounded text generation?
Here we study the balance between the number of visual instances and the number of parallel text references in the datasets, and how these two factors affect the training performance for each task.

For each task, we fix the total number of training data samples (\textit{i.e.}, unique visual-reference pairs), and set the training RPI to be $1, 2,...,n$. We have $\#sample = \#visual\_instance * RPI$.
More specifically, we train the image captioning model on MS COCO with 82,740 samples, and use 25,200 and 7,980 samples for training in the video captioning task and visual storytelling task respectively.
Figure \ref{fig:ins_ref_balance} illustrates the evaluation results for each task. For each RPI, we repeat the random sampling and training process for 10 times on each dataset.
As the training RPI increases, the performance of the image captioning model and video captioning model declines on all four metrics, while the visual storytelling performance improves. 
This suggests that introducing more visual instances during training is beneficial for the captioning tasks, where the parallel references are all objective descriptions regarding the same visual. 
In contrast, the stories in VIST are more expressive and may refer to imaginary contents~\citep{wang2018no}, leading to a much larger search space during generation. In this case, introducing more parallel references into training may help to train a more stable and better-performing storytelling model.

%===========================================================================
\section{Conclusion}
We study the sample variance in visually-grounded language generation, in terms of reference sample variance within datasets, effects of training or testing sample variance on metric scores, and the trade-off between the visual instance number and the parallel reference number per visual.
Along with some intriguing findings, we urge researchers to report sample variance in addition to the metric scores when comparing models' performance. We also recommend that when collecting a new dataset, the test set should include more parallel references for fair evaluation, while for the training set, when the text generations are expected to be distinctive and complicated, more parallel references should be collected otherwise a larger variety of visual appearances is more favorable.

\section*{Acknowledgments}
The UCSB authors were sponsored by an unrestricted gift from Google. The views and conclusions contained in this document are those of the authors and should not be interpreted as representing the sponsor.

\bibliographystyle{acl_natbib}
\bibliography{anthology,emnlp2020}

\begin{thebibliography}{18}
\expandafter\ifx\csname natexlab\endcsname\relax\def\natexlab#1{#1}\fi

\bibitem[{Anderson et~al.(2016)Anderson, Fernando, Johnson, and Gould}]{spice}
Peter Anderson, Basura Fernando, Mark Johnson, and Stephen Gould. 2016.
\newblock \href {https://doi.org/10.1007/978-3-319-46454-1_24} {Spice: Semantic
  propositional image caption evaluation}.
\newblock \emph{Lecture Notes in Computer Science}, page 382–398.

\bibitem[{Chen and Dolan(2011)}]{chen-dolan-2011-collecting}
David Chen and William Dolan. 2011.
\newblock \href {https://www.aclweb.org/anthology/P11-1020} {Collecting highly
  parallel data for paraphrase evaluation}.
\newblock In \emph{Proceedings of the 49th Annual Meeting of the Association
  for Computational Linguistics: Human Language Technologies}, pages 190--200,
  Portland, Oregon, USA. Association for Computational Linguistics.

\bibitem[{Elliott and Keller(2013)}]{meteor}
Desmond Elliott and Frank Keller. 2013.
\newblock \href {https://www.aclweb.org/anthology/D13-1128.pdf} {Image
  description using visual dependency representations}.
\newblock In \emph{Proceedings of the 2013 Conference on Empirical Methods in
  Natural Language Processing}, pages 1292--1302.

\bibitem[{Hodosh et~al.(2013)Hodosh, Young, and Hockenmaier}]{flickr8k}
Micah Hodosh, Peter Young, and Julia Hockenmaier. 2013.
\newblock \href {https://www.jair.org/index.php/jair/article/view/10833/25854}
  {Framing image description as a ranking task: Data, models and evaluation
  metrics}.
\newblock \emph{Journal of Artificial Intelligence Research}, 47:853--899.

\bibitem[{Huang et~al.(2016)Huang, Ferraro, Mostafazadeh, Misra, Agrawal,
  Devlin, Girshick, He, Kohli, Batra et~al.}]{huang2016visual}
Ting-Hao Huang, Francis Ferraro, Nasrin Mostafazadeh, Ishan Misra, Aishwarya
  Agrawal, Jacob Devlin, Ross Girshick, Xiaodong He, Pushmeet Kohli, Dhruv
  Batra, et~al. 2016.
\newblock \href {https://www.aclweb.org/anthology/N16-1147v2.pdf} {Visual
  storytelling}.
\newblock In \emph{Proceedings of the 2016 Conference of the North American
  Chapter of the Association for Computational Linguistics: Human Language
  Technologies}, pages 1233--1239.

\bibitem[{Kilickaya et~al.(2017)Kilickaya, Erdem, Ikizler-Cinbis, and
  Erdem}]{kilickaya-etal-2017-evaluating}
Mert Kilickaya, Aykut Erdem, Nazli Ikizler-Cinbis, and Erkut Erdem. 2017.
\newblock \href {https://www.aclweb.org/anthology/E17-1019} {Re-evaluating
  automatic metrics for image captioning}.
\newblock In \emph{Proceedings of the 15th Conference of the {E}uropean Chapter
  of the Association for Computational Linguistics: Volume 1, Long Papers},
  pages 199--209, Valencia, Spain. Association for Computational Linguistics.

\bibitem[{Lin(2004)}]{rouge}
Chin-Yew Lin. 2004.
\newblock \href {https://www.aclweb.org/anthology/W04-1013} {{ROUGE}: A package
  for automatic evaluation of summaries}.
\newblock In \emph{Text Summarization Branches Out}, pages 74--81, Barcelona,
  Spain. Association for Computational Linguistics.

\bibitem[{Lin et~al.(2014)Lin, Maire, Belongie, Hays, Perona, Ramanan,
  Doll{\'a}r, and Zitnick}]{coco}
Tsung-Yi Lin, Michael Maire, Serge Belongie, James Hays, Pietro Perona, Deva
  Ramanan, Piotr Doll{\'a}r, and C~Lawrence Zitnick. 2014.
\newblock \href
  {https://link.springer.com/chapter/10.1007/978-3-319-10602-1_48} {Microsoft
  coco: Common objects in context}.
\newblock In \emph{European conference on computer vision}, pages 740--755.
  Springer.

\bibitem[{Papineni et~al.(2002)Papineni, Roukos, Ward, and Zhu}]{BLEU}
Kishore Papineni, Salim Roukos, Todd Ward, and Wei-Jing Zhu. 2002.
\newblock \href {https://www.aclweb.org/anthology/P02-1040.pdf} {Bleu: a method
  for automatic evaluation of machine translation}.
\newblock In \emph{Proceedings of the 40th annual meeting on association for
  computational linguistics}, pages 311--318. Association for Computational
  Linguistics.

\bibitem[{Robertson(2004)}]{Robertson2004UnderstandingID}
S.~Robertson. 2004.
\newblock \href
  {http://citeseerx.ist.psu.edu/viewdoc/download?doi=10.1.1.438.2284&rep=rep1&type=pdf}
  {Understanding inverse document frequency: on theoretical arguments for idf}.
\newblock \emph{J. Documentation}, 60:503--520.

\bibitem[{Sharma et~al.(2017)Sharma, El~Asri, Schulz, and
  Zumer}]{sharma2017nlgeval}
Shikhar Sharma, Layla El~Asri, Hannes Schulz, and Jeremie Zumer. 2017.
\newblock \href {http://arxiv.org/abs/1706.09799} {Relevance of unsupervised
  metrics in task-oriented dialogue for evaluating natural language
  generation}.
\newblock \emph{CoRR}, abs/1706.09799.

\bibitem[{Vedantam et~al.(2015)Vedantam, Lawrence~Zitnick, and Parikh}]{cider}
Ramakrishna Vedantam, C~Lawrence~Zitnick, and Devi Parikh. 2015.
\newblock \href
  {https://www.cv-foundation.org/openaccess/content_cvpr_2015/papers/Vedantam_CIDEr_Consensus-Based_Image_2015_CVPR_paper.pdf}
  {Cider: Consensus-based image description evaluation}.
\newblock In \emph{Proceedings of the IEEE conference on computer vision and
  pattern recognition}, pages 4566--4575.

\bibitem[{Wang et~al.(2018)Wang, Chen, Wang, and Wang}]{wang2018no}
Xin Wang, Wenhu Chen, Yuan-Fang Wang, and William~Yang Wang. 2018.
\newblock \href {https://doi.org/10.18653/v1/P18-1083} {No metrics are perfect:
  Adversarial reward learning for visual storytelling}.
\newblock In \emph{Proceedings of the 56th Annual Meeting of the Association
  for Computational Linguistics (Volume 1: Long Papers)}, pages 899--909,
  Melbourne, Australia. Association for Computational Linguistics.

\bibitem[{Wang et~al.(2019)Wang, Wu, Chen, Li, Wang, and Wang}]{wang2019vatex}
Xin Wang, Jiawei Wu, Junkun Chen, Lei Li, Yuan-Fang Wang, and William~Yang
  Wang. 2019.
\newblock \href
  {https://openaccess.thecvf.com/content_ICCV_2019/papers/Wang_VaTeX_A_Large-Scale_High-Quality_Multilingual_Dataset_for_Video-and-Language_Research_ICCV_2019_paper.pdf}
  {Vatex: A large-scale, high-quality multilingual dataset for
  video-and-language research}.
\newblock In \emph{Proceedings of the IEEE International Conference on Computer
  Vision}, pages 4581--4591.

\bibitem[{Xu et~al.(2016)Xu, Mei, Yao, and Rui}]{xu2016msr}
Jun Xu, Tao Mei, Ting Yao, and Yong Rui. 2016.
\newblock \href
  {https://ieeexplore.ieee.org/stamp/stamp.jsp?tp=&arnumber=7780940} {Msr-vtt:
  A large video description dataset for bridging video and language}.
\newblock In \emph{Proceedings of the IEEE conference on computer vision and
  pattern recognition}, pages 5288--5296.

\bibitem[{Xu et~al.(2015)Xu, Ba, Kiros, Cho, Courville, Salakhudinov, Zemel,
  and Bengio}]{xu2015show}
Kelvin Xu, Jimmy Ba, Ryan Kiros, Kyunghyun Cho, Aaron Courville, Ruslan
  Salakhudinov, Rich Zemel, and Yoshua Bengio. 2015.
\newblock \href {http://proceedings.mlr.press/v37/xuc15.pdf} {Show, attend and
  tell: Neural image caption generation with visual attention}.
\newblock In \emph{International conference on machine learning}, pages
  2048--2057.

\bibitem[{Young et~al.(2014)Young, Lai, Hodosh, and Hockenmaier}]{flickr30k}
Peter Young, Alice Lai, Micah Hodosh, and Julia Hockenmaier. 2014.
\newblock \href {https://doi.org/10.1162/tacl_a_00166} {From image descriptions
  to visual denotations: New similarity metrics for semantic inference over
  event descriptions}.
\newblock \emph{Transactions of the Association for Computational Linguistics},
  2:67--78.

\bibitem[{Zhang* et~al.(2020)Zhang*, Kishore*, Wu*, Weinberger, and
  Artzi}]{bert-score}
Tianyi Zhang*, Varsha Kishore*, Felix Wu*, Kilian~Q. Weinberger, and Yoav
  Artzi. 2020.
\newblock \href {https://openreview.net/forum?id=SkeHuCVFDr} {Bertscore:
  Evaluating text generation with bert}.
\newblock In \emph{International Conference on Learning Representations}.

\end{thebibliography}

\appendix

\end{document}